\documentclass[11pt]{article}
\usepackage[hyperref]{ccl2025-en}
\usepackage{times}
\usepackage{url}
\usepackage{latexsym}
\usepackage{fancyhdr}
\usepackage{booktabs}   
\usepackage{multirow}
\usepackage{siunitx}     
\usepackage{graphicx}
\usepackage{booktabs}
\usepackage{siunitx}
\usepackage{array}   
\newcolumntype{C}[1]{>{\centering\arraybackslash}m{#1}} 
\title{System Report for CCL25-Eval Task 5:\\ New Dataset and LoRA-Fine-Tuned Qwen2.5}

\author{Haotao Xie \\
  The Hangzhou International Innovation Institute \\
  Beihang University \\
  China \\
  {\tt 1571855546@qq.com}\\}

\date{}

\begin{document}
\maketitle
\begin{abstract}
Recently, large language models (LLMs) have achieved promising progress in the fields of classical Chinese translation and the generation of classical poetry. 
However, domain-specific research on precise translation and affective-semantic understanding of classical poetry remains limited. 
The main challenge is that most studies treat the poetic appreciation task as a general-domain problem, neglecting the distinctive features of poetic appreciation, while high-quality and domain-specific datasets are extremely limited. 
To address this limitation, we decompose the task into three subtasks: term interpretation, semantic interpretation, and emotional inference. 
Based on multiple open-source datasets, we perform data cleansing and alignment to construct the Classical Chinese Poetry Instruction Pair Dataset (\textbf{CCPoetry-49K}), which comprises \textbf{49,404} high-quality instruction–response pairs explicitly optimized for this domain. 
We then propose a domain-specialized LLM, called \textbf{PoetryQwen}, by applying Low-Rank Adaptation (\textbf{LoRA}) to fine-tune the Qwen2.5-14B model. 
Experimental results on the CCL25-Eval Task 5 benchmark demonstrate that PoetryQwen achieves a score of \textbf{0.757}, representing a \textbf{9.7\%} improvement over the Qwen2.5-14B-Instruct baseline (\textbf{0.690}). 
These findings clearly indicate that PoetryQwen significantly enhances performance in precise translation and emotional understanding of classical poetry. 
We present new dataset and methodological considerations intended to support the domain-specific optimization of LLMs.
  \englishkeywords{Classical Chinese Poetry \and Instruction Dataset \and Low-Rank Adaptation \and Qwen}
\end{abstract}

\section{Introduction}
\label{intro}

%
%
\cclfootnote{
    %
    %
    \hspace{-0.65cm}  
    \textcopyright 2025 China National Conference on Computational Linguistics

    \noindent Published under Creative Commons Attribution 4.0 International License
}


Classical Chinese poetry is deeply rooted in historical and cultural contexts, often requiring knowledge of dynasties, the lives of poets, and historical events to be meaningful. The advent of LLMs offers innovative and effective solutions to this complexity, enabling more accessible understanding and lowering the barrier to learning classical poetry.
However, the majority of existing study focus on classical Chinese translation\cite{10.1145/3325887} and poetry generation\cite{zhipeng-etal-2019-jiuge,Liu_Liu_Lv_2020,zhang2024llmbasedmultiagentpoetrygeneration,yu2024charpoetchineseclassicalpoetry}, while largely overlooking the importance of systematic evaluation and interpretive assessment of classical poetry.

Furthermore, the evaluation of classical poetry is often approached as a generic NLP problem, overlooking its domain-specific challenges. This, combined with the limited availability of specialized datasets, has largely hindered the development of effective evaluation models in the field.
Therefore, it is necessary to construct domain-specific datasets and develop dedicated LLMs tailored to the unique characteristics and interpretive demands of classical Chinese poetry. Such efforts would better align with the intrinsic needs of this domain and enable more accurate, context-aware understanding and evaluation of classical poetry.
\begin{table}[htbp]
  \centering
  \begin{tabular}{@{} 
    C{2.5cm}          
    *{3}{C{3.3cm}}  
    C{1.7cm}        
    @{}}
    \toprule
    Name & 
    \shortstack{Number of Term\\ Interpretation} & 
    \shortstack{Number of Semantic\\ Interpretation} & 
    \shortstack{Number of Emotional\\ Inference} & 
    Total \\
    \midrule
    CCPoetry-49K  & 19,323 & 21,799 & 8,282 & 49,404 \\
    \bottomrule
  \end{tabular}
  \caption{Statistical Distribution of Multi-task Annotations in CCPoetry-49K} 
  \label{table1}
\end{table}
In this study, inspired by the evaluation dimensions outlined in the CCL25-Eval Task 5 benchmark\cite{chen2024largelanguagemodelsclassical}, we frame classical Chinese poetry appreciation as an evaluation-oriented task and decompose it into three subtasks. To support this, we construct CCPoetry-49K, a high-quality instruction dataset with 49K aligned samples derived from multiple open-source sources. And we apply LoRA\cite{hu2022lora} to fine-tune the Qwen2.5-14B model\footnote{\label{fn:fn2}\url{https://huggingface.co/Qwen/Qwen2.5-14B}}, resulting in PoetryQwen, a domain-adapted model specifically designed for classical poetry understanding. Experimental results show substantial gains in both translation accuracy and affective-semantic interpretation.

In summary, our main contributions are as follows:
\begin{itemize}
\item \textbf{Task Formulation:} Building upon the structure of CCL25-Eval Task 5, we adopt an evaluation framework for classical Chinese poetry that comprises three subtasks: term interpretation, semantic interpretation, and emotional inference.
\item \textbf{Dataset Construction:} We build CCPoetry-49K, a 49K-sample instruction dataset tailored to poetry evaluation, through large-scale data cleaning and alignment from open-source corpora.
\item \textbf{Model Development:} We propose PoetryQwen, a domain-specific model based on Qwen2.5-14B using LoRA fine-tuning, which achieves strong performance on classical poetry appreciation tasks.
\end{itemize}

\section{Related works}
\label{sec:related works}

\subsection{Datasets for Classical Chinese Poetry}
\label{sect:related works1}
High-quality datasets are essential for advancing the study of classical Chinese poetry using LLMs. However, existing datasets largely focus on poetry generation or translation, with limited support for multi-faceted interpretation and evaluation.

For instance, Chen \shortcite{ijcai2019p684} constructed a sentiment-labeled poetry corpus to support sentiment-controllable poetry generation, but the dataset is tailored to generation rather than comprehension or evaluation tasks. 

In the domain of classical-modern Chinese translation, Liu \shortcite{10.1145/3325887} developed a large parallel corpus that supports machine translation tasks. This work provides valuable aligned data, but lacks the interpretive and affective dimensions necessary for poetry understanding. The CCPM dataset\cite{li2021ccpmchineseclassicalpoetry} focuses on poetry matching, aligning classical poems with their modern Chinese translations, which offers a semantic bridge, yet still not cover term-level or emotional interpretation.

More recently, WenMind\cite{NEURIPS2024_5c1019b5} and WYWEB\cite{zhou2023wywebnlpevaluationbenchmark} introduced comprehensive benchmarks spanning a variety of classical Chinese tasks. WenMind includes sub-domains such as ancient prose and poetry, while WYWEB encompasses nine tasks including translation, classification, and comprehension. These benchmarks are significant steps toward domain-wide evaluation but still lack a focused dataset specifically designed for poetry appreciation tasks.

To address this gap, our work presents CCPoetry-49K, a high-quality instruction–response dataset constructed via careful alignment and cleansing from multiple open-source datasets. It uniquely supports three critical subtasks: term interpretation, semantic interpretation, and emotional inference—inspired by the evaluation structure of CCL25-Eval Task 5\footnote{\label{fn:fn1}\url{https://tianchi.aliyun.com/competition/entrance/532345}}. This dataset aims to fill the void of fine-grained, domain-specific resources for classical Chinese poetry understanding and model evaluation.

\subsection{LLMs for Classical Chinese Poetry appreciation Tasks}
\label{sect:related works2}
While recent studies have made promising progress in applying LLMs to classical Chinese poetry, their focus has primarily been on poetry generation. Systems such as Jiuge\cite{zhipeng-etal-2019-jiuge}, Deep Poetry\cite{Liu_Liu_Lv_2020}, and CharPoet\cite{yu2024charpoetchineseclassicalpoetry} enable creative generation via neural or token-free architectures, often with multimodal or user-guided input. Multi-agent generation approaches\cite{zhang2024llmbasedmultiagentpoetrygeneration} further enrich diversity and novelty.

However, poetry appreciation tasks remain underexplored. Most existing models treat poetry as a general NLP domain, overlooking its linguistic, historical, and affective complexity. Moreover, there is a notable lack of datasets and models specifically tailored for interpretive evaluation of classical poetry.

In contrast, our work focuses on evaluative understanding. Based on the structure of CCL25-Eval Task 5, we decompose the appreciation task into three subtasks and construct a high-quality instruction dataset (CCPoetry-49K). We also introduce PoetryQwen, a domain-specialized model fine-tuned via LoRA, marking a novel contribution to LLM-based poetry comprehension.

\begin{figure}
\centering
\includegraphics[
    trim=56 95 56 75, 
    clip, 
    width=1.0\textwidth
]{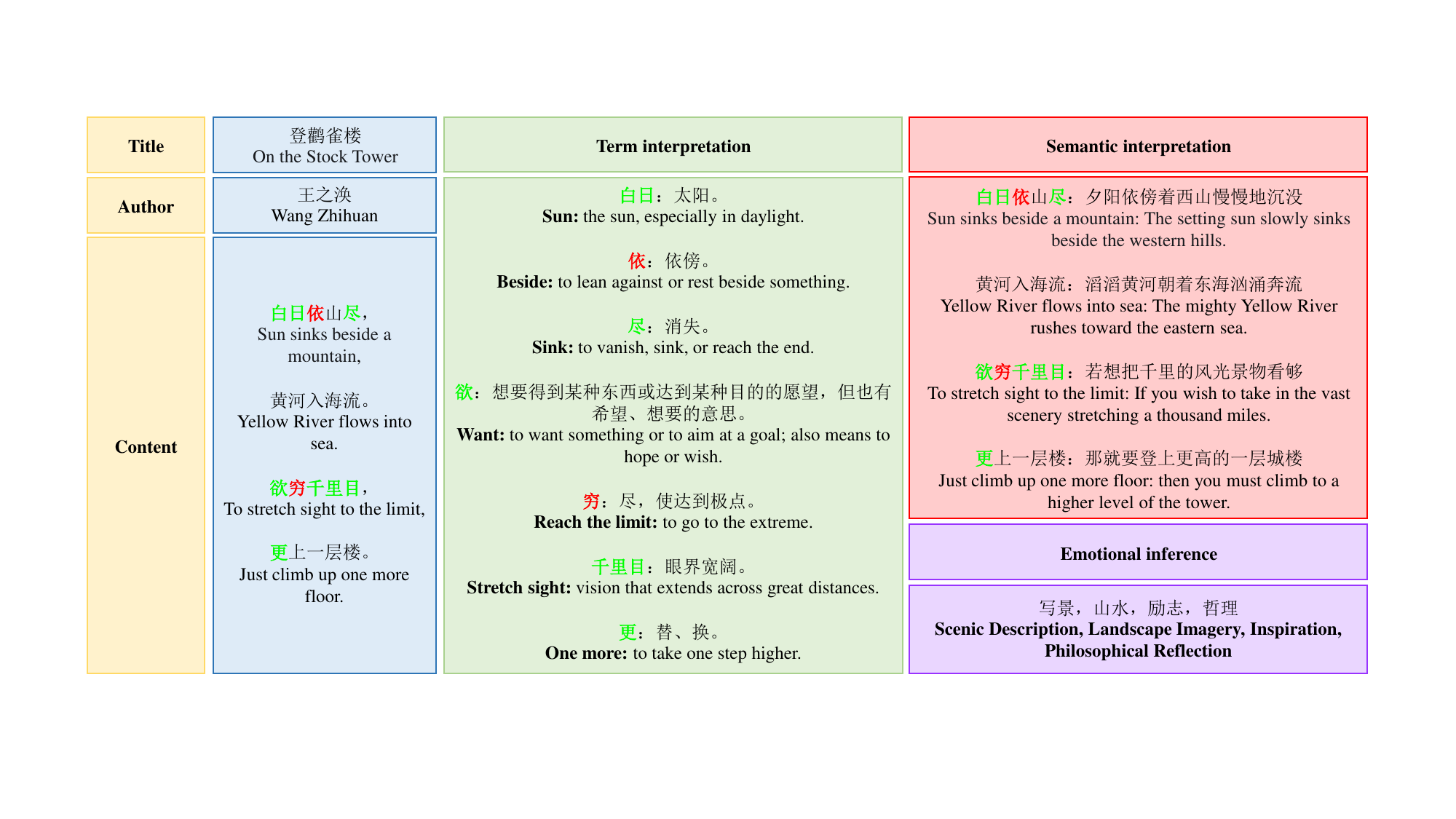}
\caption{An example extracted from multiple open-source datasets after data cleansing and alignment, including the Title, Author, Content, Term Interpretation, Semantic Interpretation, and Emotional Inference.}
\label{Fig1}
\end{figure}

\section{CCPoetry-49K Dataset}
\label{sec: CCPoetry-49K}

\subsection{Source Datasets for Poetry Appreciation}
\label{sect:Source-Datasets}
To construct a domain-specific dataset for classical Chinese poetry appreciation, we began by collecting and analyzing several open-source datasets that contain partial annotations related to poetry understanding. These datasets were primarily crawled from public educational websites and open-access online platforms, and they provide information such as modern translations, cultural term explanations, and basic sentiment annotations. However, they vary significantly in structure, granularity, and annotation quality, posing challenges for direct use in instruction tuning.

The primary datasets utilized in this study include:

\textbf{Poetry CN\footnote{\label{fn:fn3}\url{https://opendatalab.com/ABear/Poetry_CN}}:} The dataset\cite{he2024opendatalabempoweringgeneralartificial} is sourced from website\footnote{\label{fn:fn6}\url{https://www.gushici.net/}}, which compiles classical Chinese poetry along with translations, annotations, and commentary.

\textbf{Chinese ancient poetry translation\footnote{\label{fn:fn4}\url{https://github.com/YuRuiii/chinese-ancient-poetry-translation}}:} This dataset comprises aligned pairs of classical Chinese poetic lines and their corresponding human-authored modern Chinese translations.

\textbf{poems-db\footnote{\label{fn:fn5}\url{https://github.com/yxcs/poems-db}}:} Sourced from website\footnote{\label{fn:fn7}\url{https://www.gushici.net/}}, the dataset includes over 220,000 classical Chinese poems with annotations, commentaries, metadata on 10,000+ poets, 1,600+ poetic forms, 70+ dynasties, and nearly 200 thematic categories.


\subsection{Construction of CCPoetry-49K: A Domain-Specific Instruction Dataset}
\label{sect:Construction}
\begin{figure}
\centering
\includegraphics[
    trim=123 121 135 12, 
    clip, 
    width=0.9\textwidth
]{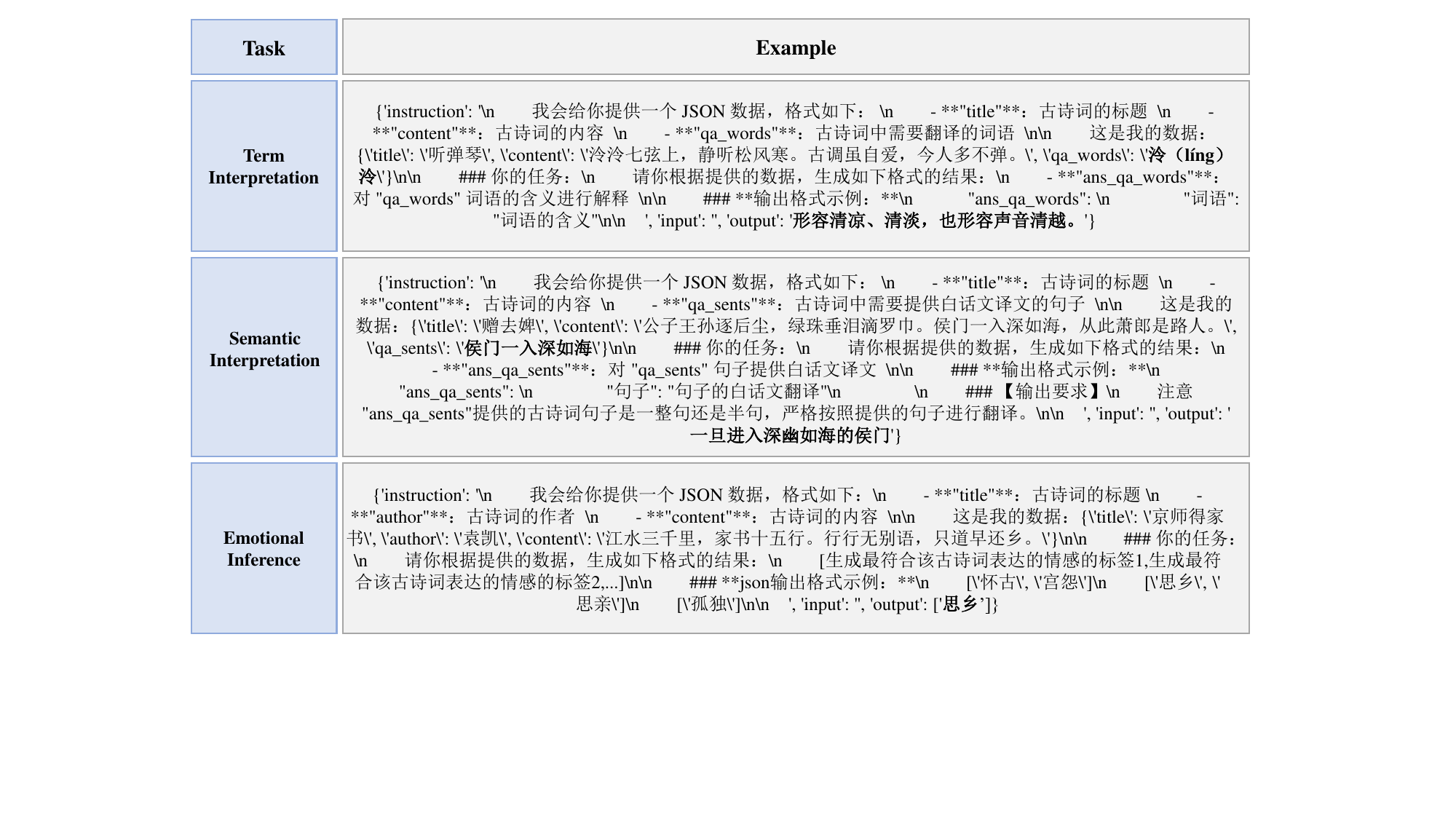}
\caption{Illustrative Subtasks Examples from the CCPoetry-49K Dataset: Term Interpretation, Semantic Interpretation, and Emotional Inference.}
\label{Fig2}
\end{figure}

Building on the open-source datasets introduced in \textbf{Section} \ref{sect:Source-Datasets}, we perform comprehensive data cleansing and alignment to obtain a high-quality foundation for downstream tasks. \textbf{Figure} \ref{Fig1} illustrates a representative example that includes key elements such as the poem’s title, author, content, term interpretation, semantic interpretation, and emotional inference. Based on this processed data, we formulate three subtasks—Term Interpretation, Semantic Interpretation, and Emotional Inference—and construct an instruction-tuning dataset tailored to each. \textbf{Figure} \ref{Fig2} provides illustrative examples for these subtasks from the proposed CCPoetry-49K Dataset, while \textbf{Table} \ref{table1} summarizes the number of instruction–response pairs per subtask, totaling 49,404 instances.

\section{Experiments}
\label{sec: Experiments}
In this section, we present the experimental setup, including model configurations, the base model used for fine-tuning, and the overall training and evaluation pipeline. We also report and analyze the experimental results to assess the effectiveness of our proposed approach.

\subsection{PoetryQwen Model}
\label{sect:PoetryQwen}
\textbf{Base Model.} We adopt Qwen2.5-14B\footnote{\label{fn:fn11}\url{https://qwenlm.github.io/blog/qwen2.5/}} as base model, with 14.7 billion parameters, which supports long-context modeling up to 128K tokens and excels in instruction following, multilingual understanding, and structured output generation.

\textbf{LoRA Fine-tuning Setup.} We apply LoRA with the following configuration: target modules include \textbf{q-proj, k-proj, v-proj, o-proj, gate-proj, up-proj} and \textbf{down-proj}; the maximum input length is set to \textbf{1240}; LoRA rank is \textbf{16}, with LoRA alpha set to \textbf{32} and dropout to \textbf{0.1}. Training is conducted for \textbf{2} epochs using a learning rate of \textbf{2e-4} and a fixed random seed of \textbf{42} to ensure reproducibility.

\begin{figure}
\centering
\includegraphics[
    trim=60 102 60 83, 
    clip, 
    width=0.8\textwidth
]{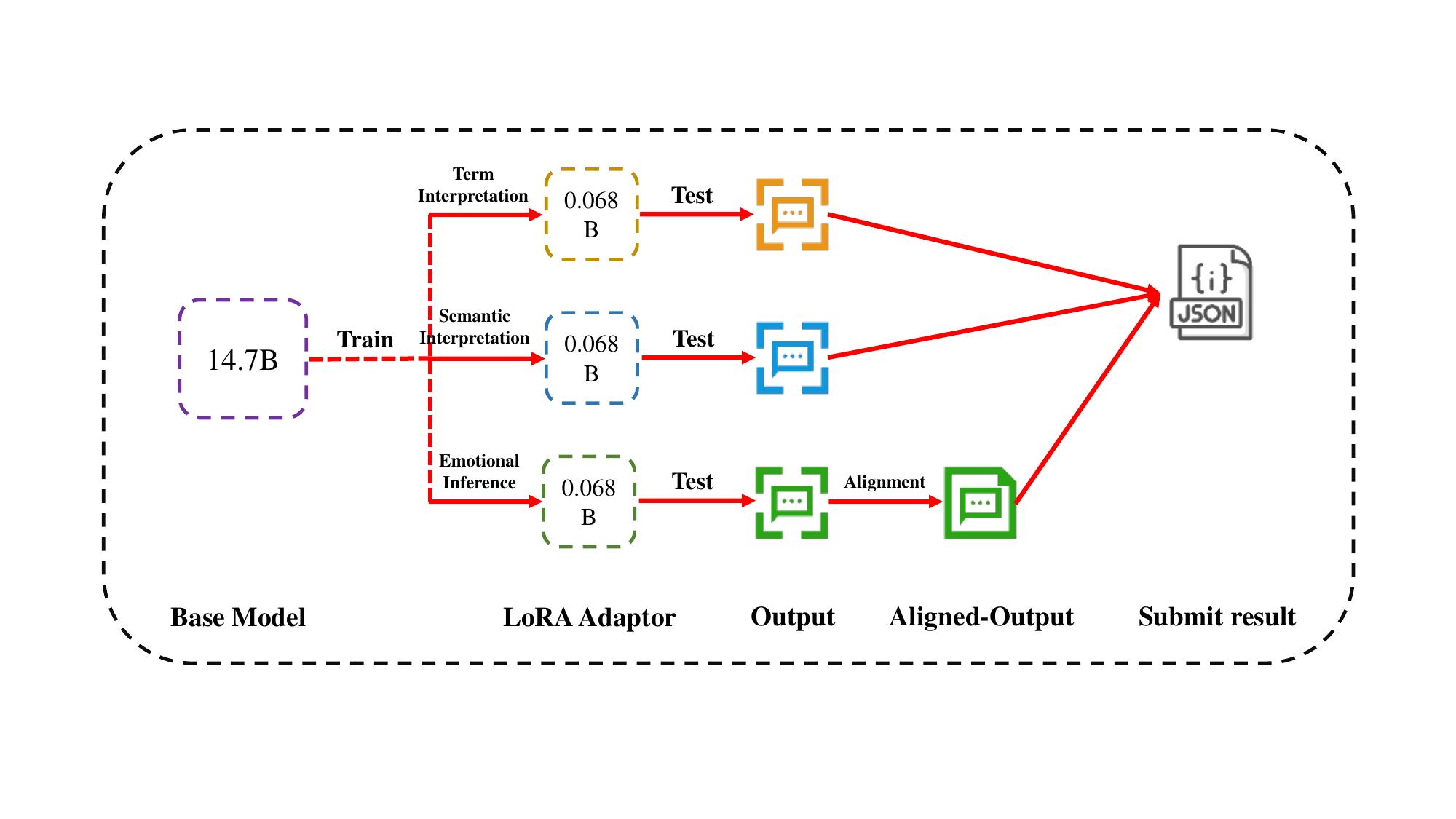}
\caption{The Structure of PoetryQwen.}
\label{Fig3}
\end{figure}

\textbf{PoetryQwen Model.} To address the three subtasks: Term Interpretation, Semantic Interpretation, and Emotional Inference, we fine-tune the Qwen2.5-14B base model using LoRA, resulting in three task-specific LoRA adapters. Each adapter is trained independently on its corresponding instruction dataset and evaluated on the respective subtask. 
Collectively, the base model and the three adapters constitute our proposed system, termed PoetryQwen, as illustrated in Figure \ref{Fig3}. For the Emotional Inference subtask, we further align the output of PoetryQwen with the submission format required by CCL25-Eval Task 5, which involves mapping the generated emotional expressions to one of the predefined emotion labels \textbf{(A, B, C, or D)}. To this end, we apply Qwen2.5-14B-Instruct for post-hoc alignment, ensuring compatibility with the evaluation requirements.

\subsection{Experiment result}
\label{sect:Experiment-result}
To evaluate the performance of our proposed dataset and model in the domain of classical Chinese poetry appreciation, we conduct experiments on the CCL25-Eval Task 5 benchmark. The evaluation encompasses three subtasks: Term Interpretation, Semantic Interpretation, and Emotional Inference with using metrics: BLEU, BERTScore, and Accuracy.

We compare our system, PoetryQwen, with several strong baselines, including Qwen2.5-7B\footnote{\label{fn:fn9}\url{https://huggingface.co/Qwen/Qwen2.5-7B}}, Qwen2.5-14B-Instruct\footnote{\label{fn:fn10}\url{https://huggingface.co/Qwen/Qwen2.5-14B-Instruct}}, and GLM-4-9B\footnote{\label{fn:fn11}\url{https://huggingface.co/THUDM/glm-4-9b}}. As shown in Table~\ref{table2}, PoetryQwen consistently outperforms all baselines across the three subtasks, demonstrating the effectiveness of our instruction-tuning strategy. These results also highlight the value of the CCPoetry-49K dataset in providing high-quality, task-specific supervision tailored for classical Chinese poetry appreciation.

\begin{table}[]
\begin{tabular}{ccccccc}
\hline
\multirow{2}{*}{Model   Name} & \multirow{2}{*}{Score} & \multicolumn{2}{c}{Term Interpretation} & \multicolumn{2}{c}{Semantic   Interpretation} & Emotional Inference \\ \cline{3-7} 
                     &       & Blue  & BertScore & Blue  & BertScore & Accuracy \\ \hline
Qwen2.5-7B           & 0.667 & 0.230 & 0.873     & 0.241 & 0.911     & 0.771    \\ \hline
Qwen2.5-14B-Instruct & 0.690 & 0.169 & 0.865     & 0.251 & 0.910     & 0.832    \\ \hline
GLM-4-9B              & 0.628 & 0.136 & 0.846     & 0.204 & 0.901     & 0.734    \\ \hline
PoetryQwen           & \textbf{0.757} & \textbf{0.405} & \textbf{0.909}     & \textbf{0.436} & \textbf{0.914}     & \textbf{0.847}    \\ \hline
\end{tabular}
\caption{Quantitative Comparison of Model Performance on Poetry Appreciation Tasks} 
\label{table2}
\end{table}


\section{Conclusion}
\label{sec:Conclusion}

In this work, we propose a domain-specific framework for classical Chinese poetry appreciation, including term interpretation, semantic interpretation, and emotional inference. 
Accordingly, we construct CCPoetry-49K, a high-quality and domain-specific instruction–response dataset with 49,404 examples. Leveraging this dataset, we fine-tune the Qwen2.5-14B model using LoRA to obtain PoetryQwen.
Experimental results on the CCL25-Eval Task 5 benchmark show a 9.7\% improvement over the base model.
Our team AI4S, registered on the Tianchi platform, demonstrates that domain specialization significantly enhances LLM performance in classical poetry understanding. Dataset will be available at https://github.com/XieHaoTao/CCPotery.

\bibliographystyle{ccl}
\bibliography{ccl2025_template/reference,ccl2025_template/reference1}

\end{document}